\documentclass[letterpaper]{article} 
\usepackage[preprint]{aaai2027}
\usepackage[hyphens]{url}  
\usepackage{graphicx} 
\usepackage{natbib}  
\usepackage{caption} 
\usepackage{algorithm}
\usepackage{algorithmicx}
\usepackage{multirow} 
\usepackage{amsmath}
\usepackage{amsfonts}
\usepackage{algpseudocode}
\usepackage{newfloat}
\usepackage{listings}
\DeclareCaptionStyle{ruled}{labelfont=normalfont,labelsep=colon,strut=off} 
\floatstyle{ruled}
\newfloat{listing}{tb}{lst}{}
\floatname{listing}{Listing}

\usepackage{subcaption}
\usepackage[table]{xcolor}
\usepackage{booktabs}
\definecolor{echorow}{RGB}{214,229,244}   
\definecolor{echosub}{RGB}{238,244,251}    
\definecolor{modelband}{RGB}{242,242,242}  
\newcommand{\best}[1]{\textbf{#1}}
\newcommand{\second}[1]{\underline{#1}}

\newcommand{\subrow}[1]{\hspace{1.1em}\textcolor{gray!75}{$\hookrightarrow$}\,#1}  

\usepackage{xspace}                        
\newcommand{\method}{\textsc{PIVOT}\xspace} 

\usepackage{hyperref}

\title{PIVOT: Efficient Query-Group Indexing for Token-Level Sparse Attention}

\author{
    Hong Liu\equalcontrib \quad 
    Yuan Cheng\equalcontrib \quad 
    Lin Niu\equalcontrib \quad 
    Yi Su \quad
    Yufei Xue \quad \\ 
    Anmin Liu \quad  
    Guanghua Yu \corresponding \quad  
    Jianchen Zhu
}

\affiliations{
    Tencent \\
    {\normalfont\small
    \texttt{lucayu@tencent.com, irisxliu@tencent.com}
    }
}

\usepackage{fancyhdr}
\usepackage{graphicx}
\usepackage{datetime}  %

\fancypagestyle{firstpagestyle}{%
  \fancyhf{}  
  \fancyhead[L]{\includegraphics[height=0.66cm]{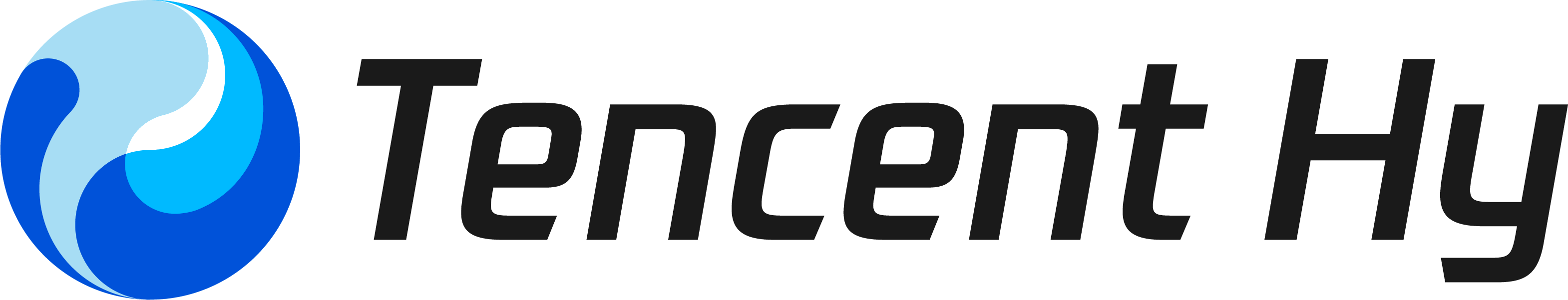}}

}

\begin{document}
\maketitle
\thispagestyle{firstpagestyle}

\begin{abstract}
Token-level sparse attention, as implemented by DeepSeek Sparse Attention (DSA) in production systems, makes the downstream attention efficient but shifts the bottleneck to the indexer that feeds it. To select the top-$k$ tokens for each query, the indexer must still score every preceding token, incurring a cost of $O(L^2)$ per layer for a sequence of length $L$. We observe that this per-query scan is largely redundant: nearby queries select highly overlapping top-$k$ tokens, and the indexer scores are long-tailed along the key axis. We exploit these properties in \textbf{\method}, \textbf{\underline{P}}roxy \textbf{\underline{I}}ndexing \textbf{\underline{V}}ia \textbf{\underline{O}}ne full-prefix \textbf{\underline{T}}raversal, a training-free, drop-in replacement for the DSA indexer that shares one prefix scan across a group of nearby queries. \method aggregates a group into a single proxy query, performs one shared full-prefix scan to obtain a candidate set, and then selects a top-$k$ for each query from that set. Two variants trade speed for fidelity: \method-Reuse shares the proxy top-$k$ across the group for maximum speed, whereas \method-Refine re-scores the candidate set with the indexer of each query and then selects an individual top-$k$, matching the dense indexer at a small additional cost. A single algorithm covers both inference phases, differing only in how groups are formed: fixed-size groups of consecutive queries in prefill, and the queries decoded together in one multi-token prediction (MTP) step in decode. On DeepSeek-V3.2 and GLM-5.1 across LongBench and RULER, \method matches the accuracy of the dense DSA indexer while accelerating it by up to $4\times$ and reducing end-to-end latency by up to $1.6\times$ at long context.


\end{abstract}

\section{Introduction}



Long-context inference has become a central challenge for LLM serving as context windows grow toward millions of tokens~\cite{dettmers2022gpt3,touvron2023llama2,yang2025qwen3,team2025kimi,xiaomi2025mimo}. At this scale, the quadratic cost of full attention dominates both latency and memory~\cite{keles2023computational}. DeepSeek Sparse Attention (DSA)~\citep{deepseekv32} addresses this by allowing each query to attend to only a small subset of the $k$ most relevant tokens rather than the entire prefix, reducing the main attention cost from $O(L^2)$ to $O(Lk)$ for a sequence of length $L$ with $k\ll L$. Concretely, DSA employs a lightweight indexer to score every preceding token for each query and select the $k$ highest-scoring ones, which are then passed to a downstream Sparse Multi-head Latent Attention (Sparse MLA) operator. This design supports production-scale models such as DeepSeek-V3.2~\cite{deepseekv32} and GLM-5.1~\cite{glm5}.


This efficiency, however, shifts the cost rather than eliminating it. Although the downstream Sparse MLA operator becomes cheap, the indexer that feeds it remains dense: to determine which tokens matter, every query must still be scored against the entire prefix. Summed over an $L$-token sequence, these per-query prefix scores yield an $O(L^2)$ indexing cost per layer. As context grows, the indexer can therefore turns from a negligible preprocessing step into the dominant cost, accounting for about $81\%$ of end-to-end latency in prefill and $41\%$ in decode at $200$K tokens~\cite{indexcache}.




\begin{figure}[!t]
    \centering
    \includegraphics[width=0.9\linewidth]{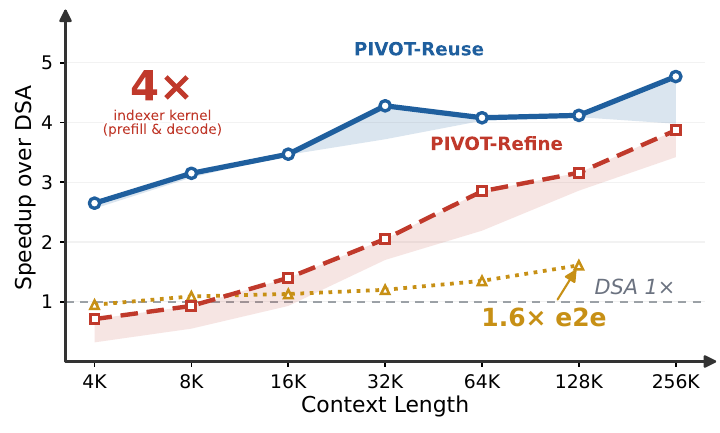}
    \vspace{-0.5cm}
    \caption{\method - faster long-context indexing}
    \vspace{-0.5cm}
    \label{fig:placeholder}
\end{figure}

\begin{figure*}[t]
    \vspace{-0.5cm}
    \centering
    \includegraphics[width=1.0\linewidth]{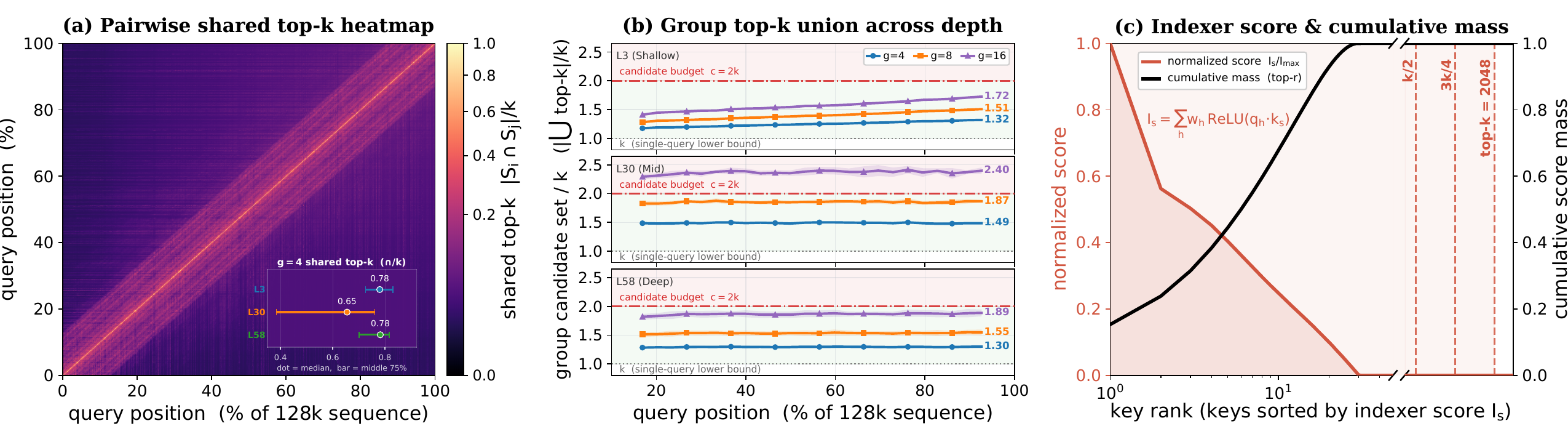}
    \caption{\textbf{Observational study on the redundancy and sparsity of indexer top-$k$ selection}. DeepSeek-V3.2, RULER-QA 128K, deep queries
    ($k_e\!>\!20$K), budget $k\!=\!2048$. \textbf{(a)~Local.} The shared-key fraction
    $|S_i\!\cap\!S_j|/k$ forms a bright diagonal: adjacent queries share $\approx\!0.8$--$0.9$ of
    their top-$k$ from shallow to deep layers(L3/L30/L58), and $0.6$--$0.8$ even across a full $g\!=\!4$ group.
    \textbf{(b)~Group-shareable.} A group's top-$k$ union stays near the candidate budget
    $c\!=\!2k$ over the 128K context ($\approx\!1.3$--$1.9\,k$ for $g\!\le\!8$).
    \textbf{(c)~Sparse.} Indexer scores $I_s\!=\!\sum_h w_h\,\mathrm{ReLU}(q_h\!\cdot\!k_s)$
    concentrate on a small active set, saturating far below $k/2,3k/4,k$.}
    \label{fig:obs}
    \vspace{-0.5cm}
\end{figure*}

Recent work has reduced this indexer cost from several angles: lowering the precision of each score, or reducing indexer computation along the token axis (HISA~\citep{hisa}), the head axis (MISA~\citep{misa}), or the layer axis (IndexCache~\citep{indexcache}). These methods, however, leave the query axis untouched: the indexer is still invoked separately for each query, and each call still scans the full prefix. We instead examine the \emph{query axis} and find this per-query search largely redundant. Figure~\ref{fig:obs}a plots the pairwise top-$k$ overlap between queries: nearby queries lie on a bright near-diagonal band, meaning they select almost the same tokens. Figure~\ref{fig:obs}b further shows that, for a group of $g$ queries, the union of their top-$k$ sets remains close to $k$ and far below its worst case of $g\,k$. Thus, a candidate set only slightly larger than $k$ suffices to cover the whole group. This redundancy suggests a new axis for reducing indexer cost: sharing one full-prefix scan across a group of nearby queries instead of repeating it for each.

Motivated by this observation, we propose \method (\emph{Proxy Indexing Via One full-prefix Traversal}), a training-free, drop-in replacement for the DSA indexer that shares one prefix scan across a group of nearby queries instead of running one per query. \method proceeds in two steps. First, it aggregates a group into a single proxy query, scans the prefix once, and retains the top-$c$ scoring tokens as a shared candidate set $\mathcal{C}$. As Figure~\ref{fig:obs} shows, nearby queries select heavily overlapping top-$k$ tokens and the indexer scores are long-tailed, so a budget only slightly above $k$ and far below the sequence length $L$ suffices ($k<c\ll L$) for $\mathcal{C}$ to cover most tokens needed by the group. Second, each query obtains its top-$k$ through one of two varians: \method-Refine re-scores $\mathcal{C}$ with the exact indexer of each query and selects an individual top-$k$, whereas the cheaper \method-Reuse skips this step and assigns the proxy top-$k$ to every query in the group. A group of $g$ queries therefore replaces $g$ full-prefix scans with one shared scan and a lightweight per-query step, reducing the group indexing cost from $\mathcal{O}(gL)$ to $\mathcal{O}(L+gc)$ for Refine or $\mathcal{O}(L)$ for Reuse, while the exact re-scoring in Refine preserves accuracy. The resulting top-$k$ indices are passed to Sparse MLA through the same interface as in DSA, so the downstream operator remains unchanged. A single algorithm serves both phases, differing only in how groups are formed: prefill partitions the simultaneously available query positions into fixed-size groups of $g$, whereas decode takes the $d{+}1$ tokens of one multi-token-prediction (MTP)~\citep{glm5} step as a group. \method thus rides on top of MTP at no additional cost, and the two speedups compound.

\method saves indexing cost entirely along the query axis. We summarize our contributions as follows:

\begin{itemize}
\item \textbf{A new efficiency axis.}
We identify substantial cross-query redundancy in DSA indexing: nearby queries select highly overlapping top-$k$ tokens. This redundancy defines a fourth efficiency axis for accelerating token-level sparse attention, which is orthogonal to prior methods that operate along the token, head, and layer axes.

\item \textbf{\method.}
We introduce \method, a training-free, drop-in indexer that preserves the DSA interface while amortizing one full-prefix scan over a group of queries. \method provides two variants: \method-Refine re-scores each query within a small shared candidate set to obtain its top-$k$, whereas \method-Reuse shares the proxy top-$k$ across the group. This reduces the per-group indexing cost from $\mathcal{O}(gL)$ toward $\mathcal{O}(L)$, and the same algorithm serves both inference phases by using fixed-size groups in prefill and same-step MTP groups in decode.

\item \textbf{Results.} On DeepSeek-V3.2 and GLM-5.1 across LongBench and RULER, \method matches the accuracy of the dense DSA indexer while accelerating the indexer operator by up to $4\times$ and lowering end-to-end latency by up to $1.6\times$ at long context.

\end{itemize}

\section{Related Work}
\begin{figure*}[htbp]
\vspace{-0.5cm}
    \centering
    \includegraphics[width=1.0\linewidth]{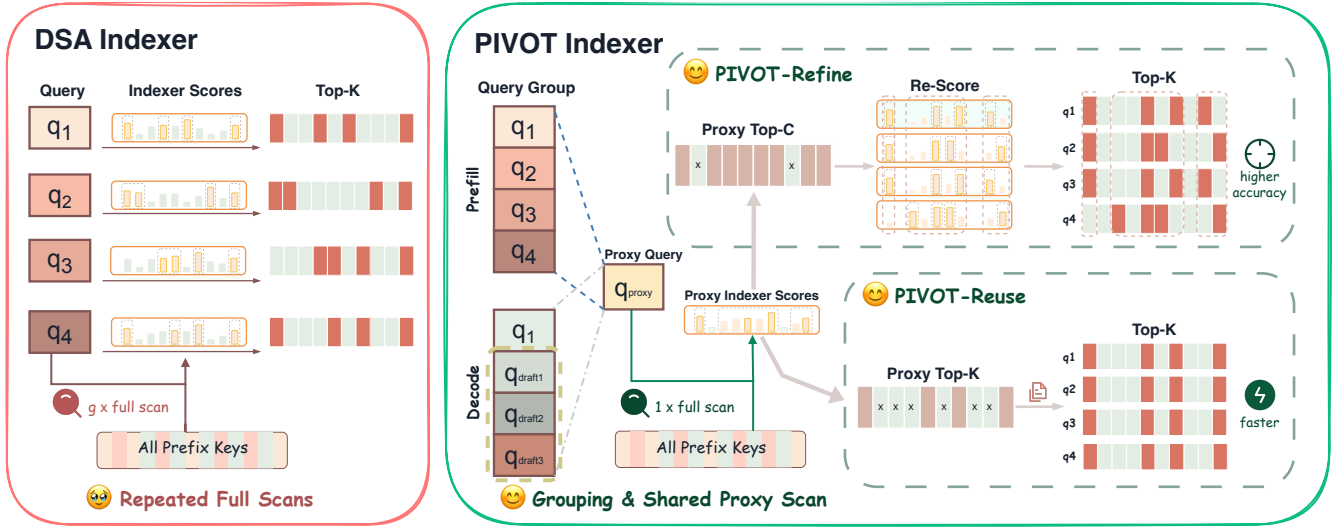}
\caption{\textbf{\method amortizes the DSA indexer's per-query prefix scans across a query group.}
\textbf{Left (DSA):} To select its top-$k$, every query $q_1,\dots,q_g$ independently scores \emph{all} prefix keys, so a group of $g$ queries incurs $g$ full-prefix scans.
\textbf{Right (\method):} The queries of a group (fixed-size in \emph{prefill}; the current token together with its MTP draft tokens in \emph{decode}) are aggregated into a single \emph{proxy query} $q_{\mathrm{proxy}}$ which performs one shared full-prefix scan to produce proxy indexer scores. Two variants then select each query's $k$ highest-scoring key tokens as its top-$k$: \textbf{\method-Refine} keeps the proxy's top-$c$ candidates and re-scores them per query for higher accuracy, whereas \textbf{\method-Reuse} directly reuses the proxy's top-$k$ for the whole group for higher speed. }
    \label{fig:method}
    \vspace{-0.5cm}
\end{figure*}

\subsubsection{Sparse attention.}

Sparse attention reduces the cost of long-context inference by allowing each query to attend to only a subset of tokens. Block-level methods estimate the importance of contiguous key blocks either online~\citep{jiang2024minference,lai2025flexprefill,stem,wang2025proxyattn,xu2025xattention,tang2024quest} or through learned block routing during training, as in MoBA~\citep{lu2502moba}, InfLLM-V2~\citep{zhao2025infllm}, and NSA~\citep{yuan2025nsa}. Other approaches induce block-sparse attention with lightweight training, including SeerAttention~\citep{gao2024seerattention,gao2025seerattention}, DuoAttention~\citep{xiao2025duoattention}, and RTPurbo~\citep{zhou2026full}. These block-level designs are hardware-friendly but coarse, because all tokens in a block are kept or dropped together. Token-level methods, in contrast, score individual tokens for finer selection. DSA~\citep{deepseekv32} is representative: a lightweight indexer scores every prefix token for each query and selects the top-$k$. This finer granularity shifts the bottleneck to the indexer, which must still scan the full prefix for every query in long-context settings. \method preserves the token-level selection of DSA while amortizing these redundant per-query scans.

\subsubsection{Accelerating the DSA indexer.}
A growing line of work accelerates the DSA indexer itself. HISA~\citep{hisa} replaces the flat token scan with a block-to-token hierarchy, in which each query first filters candidate blocks and then refines the selection within them. MISA~\citep{misa} treats indexer heads as a mixture of experts and activates only a query-dependent subset. IndexCache~\citep{indexcache} exploits cross-layer redundancy by reusing the top-$k$ set of one layer in nearby layers. SparDA~\citep{fu2026sparda} uses a lightweight predictor to prefetch next-layer KV blocks. These methods reduce indexer cost along the token, head, and layer axes by making individual indexer calls cheaper or by skipping some layer-wise calls. However, they leave the query axis untouched: the indexer is still invoked separately for each query over the full prefix.

\subsubsection{Sharing computation across queries.}
Cross-query sharing has appeared primarily in token pruning and KV-cache compression. LazyLLM~\citep{fu2024lazyllm} progressively prunes prefix tokens during prefill, so tokens dropped for earlier queries remain unavailable to later ones. H2O~\citep{zhang2024h2o} and TOVA~\citep{oren2024tova} evict low-scoring entries from the KV cache based on accumulated or recent attention scores, and SnapKV~\citep{li2024snapkv} uses an observation window to select important KV positions that later decoding steps reuse. In each case, sharing amounts to permanently discarding tokens. \method differs in this respect: it discards no token, the downstream attention still sees the full prefix, and only the index-selection step is shared.

\section{Preliminary}
\label{sec:preliminary}
\subsection{Background}
We build on DSA~\citep{deepseekv32}, a token-level sparse attention mechanism with two components: a \emph{lightning indexer} that selects the preceding tokens each query should attend to, and a Sparse MLA operator that attends only to the selected tokens. For a query at position $t$ and a preceding token at position $s$, the indexer estimates their relevance with $H^I$ lightweight heads,
\vspace{-0.2cm}
\begin{equation}
I_{t,s} = \sum_{j=1}^{H^I} w^{I}_{t,j}\,\mathrm{ReLU}\!\big(\mathbf{q}^{I}_{t,j}\cdot \mathbf{k}^{I}_{s}\big),
\label{eq:indexer}
\end{equation}
where the indexing query $\mathbf{q}^{I}_{t,j}$ and gating weight $w^{I}_{t,j}$ are projected from token $t$, the indexing key $\mathbf{k}^{I}_{s}$ is projected from token $s$, and $H^I$ denotes the number of indexer heads. The indexer retains the $k$ highest-scoring tokens as the index set $\mathcal{T}_t=\mathrm{TopK}(I_{t,:},\, k)$, over which Sparse MLA attends. This reduces the main attention cost from $O(L^2)$ to $O(Lk)$ for a sequence of length $L$ with $k\ll L$. Thus, $\mathcal{T}_t$ is the sole interface between the indexer and Sparse MLA, and \method only changes how $\mathcal{T}_t$ is produced, leaving Sparse MLA and the KV cache unchanged.

\subsubsection{Cost.} Sparse MLA is now efficient, but producing $\mathcal{T}_t$ is not: by \eqref{eq:indexer}, every query must still be scored against the entire prefix, at $O(L^2)$ per layer in prefill and $O(L)$ per step in decode. As context grows, the indexer therefore turns from a negligible preprocessing step into the dominant cost, accounting for about $81\%$ of end-to-end latency in prefill and $41\%$ in decode at $200$K tokens~\citep{indexcache}. This per-query full-prefix scan is exactly the cost that \method reduces.

\subsection{Observations}
The DSA indexer runs an independent full-prefix scan for every query, which is what makes it expensive at long context. To determine whether this per-query cost is truly necessary, we examine how the indexer behaves on DeepSeek-V3.2 at $128$K, and identify three properties, summarized in Figure~\ref{fig:obs}.

\subsubsection{O1: neighboring queries overlap heavily.}
Neighbouring queries select highly similar token sets. Figure~\ref{fig:obs}a plots the pairwise shared top-$k$ fraction $|\mathcal{T}_i\cap\mathcal{T}_j|/k$: a bright band hugs the diagonal, and a group of four shares a median of $65$--$78\%$ of its top-$k$ from shallow to deep layers (L3/L30/L58), lowest in the middle layers but still a clear majority. The band fades only slowly off the diagonal, so queries several positions apart still agree on most of their selection. This property follows from locality: adjacent queries share almost the same prefix and carry similar hidden states, hence similar indexing queries, so the keys they score as relevant drift gradually rather than jump.


\subsubsection{O2: a whole group's top-$k$ has a small union.}
Across a whole group, the combined top-$k$ can in principle range from $k$ tokens, when all members select identically, to $g\,k$, when they select disjointly. Figure~\ref{fig:obs}b shows that it sits near the low end of this range. Averaged over depths, the union $|\bigcup_i \mathcal{T}_i|/k$ over a group of $g$ queries measures only $1.3$--$1.5\times k$ at $g{=}4$, and it grows sub-linearly, far below the worst-case slope of $g$, reaching just $1.7$--$2.4\times k$ at $g{=}16$. The group therefore concentrates on a shared core and adds only a thin, slowly widening margin of query-specific tokens.


\subsubsection{O3: indexer scores are long-tailed.}
The index score in \eqref{eq:indexer} is heavily long-tailed along the key axis, as Figure~\ref{fig:obs}c shows. The per-head ReLU is the cause: each dot product $\mathbf{q}^{I}_{t,j}\!\cdot\!\mathbf{k}^{I}_{s}$ is near zero-mean, ReLU zeros out the negative interactions, and only keys that align positively across many heads accumulate a large score. This produces a sharp separation between a few dominant keys and a bulk that collapses toward zero: the cumulative-mass curve rises steeply and then flattens, so a small fraction of keys already carries most of the total score mass. This long-tailed shape holds across depths, although its steepness varies from layer to layer.

Taken together, these three properties motivate a simple design. Since neighbouring queries select highly overlapping top-$k$ tokens (O1), a single \emph{proxy} query can represent an entire group, replacing $g$ separate prefix scans with one shared scan. Because the union of the group selections remains small (O2), the tokens needed by the group members fit within a shared candidate set of size $c\!\approx\!2k\ll L$, far below the worst case of $g\cdot k$. Because indexer scores are long-tailed (O3), re-scoring each query within this small set recovers nearly the same top-$k$ as a full-prefix search. These three properties play complementary roles: O1 enables sharing, O2 makes it efficient, and O3 preserves accuracy.

Motivated by these properties, we propose \method, a training-free, drop-in replacement for the DSA indexer that amortizes prefix scans across nearby queries. \method aggregates each group into a proxy query, performs one shared scan to obtain a candidate set $\mathcal{C}$ of size $c$, and then selects a top-$k$ for each query from $\mathcal{C}$.




\begin{figure*}[htbp]
    \centering
    \includegraphics[width=1.0\linewidth]{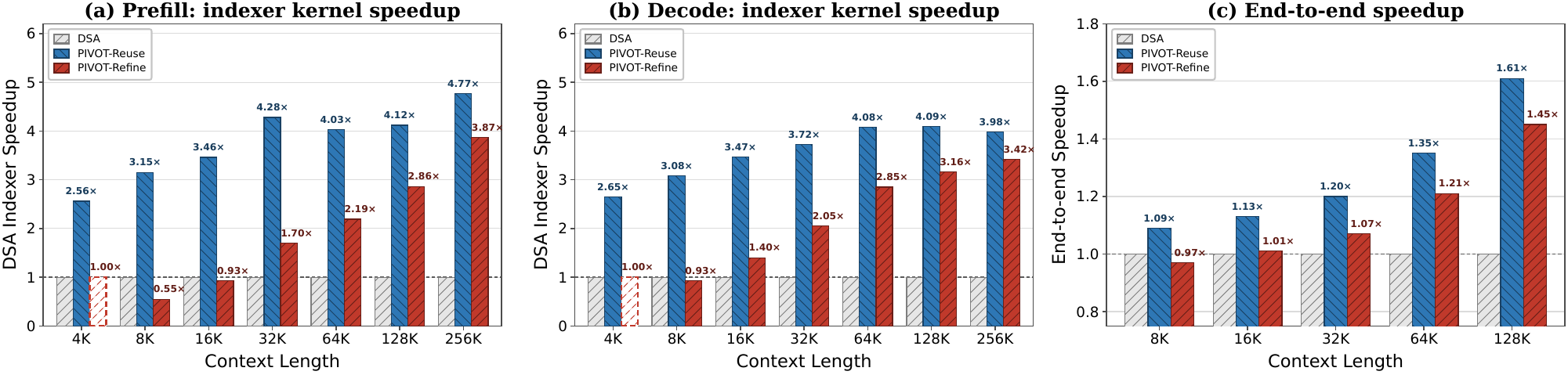}
\caption{\textbf{Indexer speedup over the DSA baseline on DeepSeek-V3.2}, from
4K to 256K. \textbf{(a)} Prefill and \textbf{(b)} decode report the
indexer-kernel speedup, while \textbf{(c)} reports the end-to-end speedup, with
the DSA baseline normalized to $1\times$. \method-Refine falls back to DSA at 4K (dashed, $1\times$).}
    \label{fig:speedup}
    \vspace{-0.5cm}
\end{figure*}

\section{Method}
\label{sec:method}

\method replaces the DSA indexer with a drop-in module that exposes the same interface, the per-query index set $\mathcal{T}_t$. Instead of running a full-prefix traversal for every query, it amortizes one scan over a \emph{group} of nearby queries, which is sound because neighbouring queries select highly overlapping tokens (O1). The group is aggregated into a single \emph{proxy query} that scans the prefix once and scores every token (the coarse step), and each query then derives its top-$k$ from these shared scores (the fine step). This yields two variants, in increasing order of fidelity. \method-Reuse skips the fine step and shares the proxy top-$k$ across the group. \method-Refine keeps a top-$c$ candidate set from the coarse scores and re-scores it per query. Here O2 allows the budget to stay small ($k<c\ll L$), and O3 makes the re-ranking reliable. Both variants keep $\mathcal{T}_t$ as the sole interface, leaving Sparse MLA and the KV cache unchanged.
The same algorithm applies to both inference phases, which differ only in how a group is formed. Figure~\ref{fig:method} contrasts \method with the per-query DSA indexer and shows its shared proxy scan and two variants.


\subsection{Grouping}
\label{sec:grouping}
\method groups queries that are available simultaneously, so one proxy scan is amortized across them within a single step, with no state carried across steps. 

\subsubsection{Prefill.} All query positions of a sequence are available at once, so \method partitions them into contiguous groups of a fixed size $g$. A group starting at position $t$ is
\begin{equation}
G^{\mathrm{P}}_{t} = \{\,q_{t},\ q_{t+1},\ \dots,\ q_{t+g-1}\,\},
\label{eq:prefill-group}
\end{equation}
where $q_t$ is the query at position $t$. A group never crosses a request boundary, and the last group of a request may be smaller. Partitioning the $L$ positions of a layer into groups of size $g$ reduces the indexing cost from $\mathcal{O}(L^2)$ to $\mathcal{O}(L^2/g)$ up to the fine step, which is the dominant saving at long context.

\subsubsection{Decode.}
Decoding generates one token at a time, leaving no natural query group. \method obtains one for free from MTP, a speculative decoding scheme used in production-scale models such as DeepSeek-V3.2 and GLM-5.1. At step $t$, the target model produces the current query $q_t$, and an MTP draft head, conditioned on the target hidden state, autoregressively proposes $d$ future queries $\tilde{q}_{t+1},\dots,\tilde{q}_{t+d}$. Since these queries are already evaluated within the same decoding step, \method takes them as one group,
\begin{equation}
G^{\mathrm{D}}_{t} = \{\,q_t,\ \tilde{q}_{t+1},\ \dots,\ \tilde{q}_{t+d}\,\},\qquad g = d + 1 .
\label{eq:decode-group}
\end{equation}
\method thereby reuses the query batch already formed by MTP at no additional cost ($d{=}3$, hence $g{=}4$, in both models). \method thus complements MTP rather than competing with it, and the two speedups compound.

\subsection{The Shared Proxy Scan}
\label{sec:proxy}

Given a group $G$ (from either phase), \method runs the coarse step once for the entire group. It first aggregates the group into a single \emph{proxy query}, whose indexing query and gating weight are the per-head averages over the members of $G$,
\begin{equation}
\bar{\mathbf{q}}^{I}_{j} = \frac{1}{g}\sum_{t'\in G}\mathbf{q}^{I}_{t',j},
\qquad
\bar{w}^{I}_{j} = \frac{1}{g}\sum_{t'\in G} w^{I}_{t',j},
\label{eq:proxy}
\end{equation}
for each indexer head $j=1,\dots,H^I$. The proxy then scores the entire prefix once through the indexer function in Eq.~\eqref{eq:indexer},
\vspace{-0.2cm}
\begin{equation}
\bar{I}_{s} = \sum_{j=1}^{H^I} \bar{w}^{I}_{j}\,\mathrm{ReLU}\!\big(\bar{\mathbf{q}}^{I}_{j}\cdot \mathbf{k}^{I}_{s}\big),\qquad s\le t,
\label{eq:proxy-score}
\end{equation}
where the causal cutoff at the first position of the group, $t$, keeps the scan valid for every member of $G$. This single scan, at a cost of $\mathcal{O}(L)$, is shared by the group. The two variants below build on the proxy scores $\{\bar{I}_{s}\}$ without scanning the prefix again: \method-Reuse directly takes their top-$k$, while \method-Refine forms a candidate set and re-scores it per query. We use mean pooling for the proxy, as it works best among the aggregations we examined.

\subsection{PIVOT-Reuse}
\label{sec:reuse}


The simplest option is to skip the fine step and let the proxy selection serve the entire group. \method-Reuse takes the top-$k$ of the proxy scores over the prefix and assigns it to every query in the group,
\begin{equation}
\mathcal{T}_t = \mathrm{TopK}\big(\{\,\bar{I}_{s}\mid s\le t\,\},\, k\big)
\qquad\text{for all } t\in G .
\label{eq:reuse}
\end{equation}
Because neighbouring queries select highly overlapping sets of tokens (O1), the proxy top-$k$ is already a close approximation of what each member of the group would select on its own. This variant performs no per-query scoring: the group is served entirely by the shared scan of \eqref{eq:proxy-score} together with a single top-$k$ selection, giving an indexing cost of $\mathcal{O}(L)$ against $\mathcal{O}(gL)$ for DSA. Reuse is therefore the faster variant. A single shared selection, however, cannot capture query-specific differences. Although small under O1, these differences are nonzero and can cost some accuracy at very long context, which motivates the refinement of \method-Refine.

\subsection{\method-Refine}
\label{sec:refine}

\method-Refine preserves the shared scan while restoring the per-query specificity discarded by Reuse. It first forms a shared candidate set from the proxy top-$c$ scores,
\begin{equation}
\mathcal{C} = \mathrm{TopK}\big(\{\,\bar{I}_{s}\mid s\le t\,\},\, c\big),
\label{eq:candidate}
\end{equation}
where the budget $c$ exceeds $k$ because $\mathcal{C}$ must cover the \emph{union} of the group's top-$k$ sets, not just the $k$ tokens of a single query. How large $c$ must be is governed by O2: since this union stays close to $k$, a small budget already suffices, and we use $c=2k$ throughout ($k<c\ll L$). Each query then re-scores these candidates with \emph{its own} indexing query and gate, applying the exact indexer of \eqref{eq:indexer} over $\mathcal{C}$,
\vspace{-0.2cm}
\begin{equation}
I_{t,s} = \sum_{j=1}^{H^I} w^{I}_{t,j}\,\mathrm{ReLU}\!\big(\mathbf{q}^{I}_{t,j}\cdot \mathbf{k}^{I}_{s}\big),\quad s\in\mathcal{C},
\label{eq:refine-score}
\end{equation}
and selects its own top-$k$ from these query-specific scores,
\begin{equation}
\mathcal{T}_t = \mathrm{TopK}\big(\{\,I_{t,s}\mid s\in\mathcal{C}\,\},\, k\big).
\label{eq:refine}
\end{equation}
Because the exact scoring runs over the $c$ candidates rather than the full prefix, refinement adds only $\mathcal{O}(gc)$ per group on top of the shared scan. With $c\ll L$ this overhead is small, and a group costs $\mathcal{O}(L+gc)$, which lies between $\mathcal{O}(L)$ for Reuse and $\mathcal{O}(gL)$ for DSA. Refinement also preserves accuracy well: because the indexer scores are long-tailed (O3), most of the truly high-scoring tokens for each query already lie within the shared candidate set $\mathcal{C}$, so re-ranking $\mathcal{C}$ closely approximates the top-$k$ that a full-prefix scan would return. In practice, we take the union of $\mathcal{C}$ with a small per-query local window so that each query retains its most recent tokens, see Appendix for details.

\definecolor{tblmodel}{HTML}{DCE6F1}
\definecolor{tblecho}{HTML}{E8F1FA}
\definecolor{tblechosub}{HTML}{F3F7FC}
\providecommand{\best}[1]{\textbf{#1}}
\providecommand{\second}[1]{\underline{#1}}
\providecommand{\subrow}[1]{\hspace{0.7em}\(\hookrightarrow\)\,#1}

\begin{table*}[t]
\centering

\begingroup
\small
\setlength{\tabcolsep}{8pt}
\renewcommand{\arraystretch}{1.0}

\resizebox{\textwidth}{!}{%
\begin{tabular}{l *{7}{c} !{\color{black!35}\vrule width 0.4pt} *{7}{c}}

\toprule
& \multicolumn{7}{c}{\textbf{LongBench}}
& \multicolumn{7}{c}{\textbf{RULER}} \\
\cmidrule(lr){2-8}\cmidrule(lr){9-15}
\textbf{Method}
& \textbf{SQA} & \textbf{MQA} & \textbf{Sum}
& \textbf{FS} & \textbf{Syn} & \textbf{Code} & \textbf{AVG}
& \textbf{4K} & \textbf{8K} & \textbf{16K}
& \textbf{32K} & \textbf{64K} & \textbf{128K} & \textbf{AVG} \\
\midrule

\multicolumn{15}{c}{\rule{0pt}{2.3ex}\textbf{DeepSeek-V3.2}} \\
\midrule
DSA
& 50.88 & 53.11 & 22.32 & 64.85 & 69.67 & 74.84 & 55.95
& 96.41 & 95.71 & 96.12 & 95.77 & 91.32 & 90.45 & 94.30 \\
HISA
& 51.09 & \second{52.98} & 22.37 & 64.11
& \best{69.83} & 74.70 & 55.85
& 95.90 & 95.44 & 92.21 & 83.83 & 77.08 & 62.46 & 84.49 \\
MISA
& 51.16 & 51.93 & 22.38 & 63.97
& 69.33 & 73.80 & 55.43
& 95.90 & 94.64 & 95.29 & 94.28 & 90.50 & 84.91 & 92.59 \\
IndexCache
& \best{51.23} & 52.55 & 22.09 & 65.06
& 69.67 & \best{75.81} & 56.07
& \second{96.03} & \best{96.37} & 95.38 & \best{96.33}
& \best{92.19} & 88.67 & \second{94.16} \\

\cmidrule(lr){1-15}
\rowcolor{tblecho}
\textbf{PIVOT}-{\scriptsize\bfseries Reuse}
& 51.04 & 52.59 & \best{22.50} & \second{65.27}
& 69.50 & 75.55 & \second{56.08}
& \second{96.03} & \second{95.86} & 94.81 & 95.16
& \second{90.97} & 86.46 & 93.22 \\
\rowcolor{tblechosub}
\subrow{+ IC}
& \second{51.18} & \best{53.02} & \second{22.43}
& 64.44 & 69.61 & 75.60 & 56.05
& \best{96.41} & 95.80 & \best{96.12} & 95.37
& 88.47 & 85.23 & 92.90 \\
\rowcolor{tblecho}
\textbf{PIVOT}-{\scriptsize\bfseries Refine}
& 51.14 & 52.77 & 22.34 & \best{65.38}
& \second{69.78} & \second{75.66} & \best{56.18}
& \best{96.41} & \second{95.86} & \second{96.06}
& \second{96.02} & 90.46 & \best{90.63} & \best{94.24} \\
\rowcolor{tblechosub}
\subrow{+ IC}
& 50.97 & 52.38 & 22.33 & 64.85
& \best{69.83} & 75.37 & 55.96
& \best{96.41} & 95.15 & 95.58 & 95.85
& 90.05 & \second{89.07} & 93.69 \\

\midrule
\multicolumn{15}{c}{\rule{0pt}{2.3ex}\textbf{GLM-5.1}} \\
\midrule
DSA
& 49.40 & 56.20 & 23.80 & 66.56 & 69.04 & 75.73 & 56.79
& 95.51 & 96.15 & 96.03 & 96.03 & 95.27 & 92.14 & 95.19 \\
HISA
& 48.75 & 55.05 & 23.76 & 66.60
& \second{69.61} & 75.26 & 56.51
& \second{95.51} & \second{96.15} & 93.77 & 88.00
& 80.12 & 73.06 & 87.77 \\
MISA
& \best{49.56} & 55.03 & 23.76 & 67.14
& \best{71.96} & 75.39 & \best{57.14}
& \best{96.41} & 95.64 & \best{96.54} & 96.04
& 94.51 & 74.40 & 92.26 \\
IndexCache
& 49.00 & \second{55.78} & 23.70 & \second{67.52}
& 68.86 & 76.22 & 56.85
& 95.00 & 95.64 & 95.77 & 96.33
& \best{95.31} & \best{92.99} & \best{95.17} \\

\cmidrule(lr){1-15}
\rowcolor{tblecho}
\textbf{PIVOT}-{\scriptsize\bfseries Reuse}
& 48.92 & 55.09 & 23.76 & 67.14
& 68.84 & 75.94 & 56.62
& 94.62 & \second{96.15} & 96.06 & 95.62
& 93.59 & 88.86 & 94.15 \\
\rowcolor{tblechosub}
\subrow{+ IC}
& 49.17 & 55.41 & \best{23.86} & 66.57
& 68.80 & 75.83 & 56.61
& 95.13 & 96.03 & 95.98 & 93.06
& 92.67 & 86.28 & 93.19 \\
\rowcolor{tblecho}
\textbf{PIVOT}-{\scriptsize\bfseries Refine}
& \second{49.31} & \best{55.85} & 23.73 & 67.22
& 69.18 & \best{76.42} & \second{56.95}
& 95.42 & \second{96.15} & \second{96.15}
& \second{96.46} & \second{95.03} & 91.79 & \best{95.17} \\
\rowcolor{tblechosub}
\subrow{+ IC}
& 49.16 & 55.19 & \second{23.81} & \best{67.70}
& 69.33 & \second{76.34} & 56.92
& 94.87 & \best{96.41} & 95.31 & \best{96.50}
& 94.41 & \second{91.88} & \second{94.90} \\
\bottomrule
\end{tabular}%
}
\endgroup
\vspace{-0.2cm}
\caption{\textbf{Results (\%) on LongBench and RULER.}
We evaluate DeepSeek-V3.2 and GLM-5.1 across LongBench categories spanning
Single-Document QA (SQA), Multi-Document QA (MQA), Summarization (Sum),
Few-Shot Learning (FS), Synthetic (Syn), and Code Completion (Code), and across RULER context lengths
from 4K to 128K. AVG is the macro-average within each benchmark, and IC denotes
IndexCache. \colorbox{tblecho}{Shaded} rows indicate our methods;
\textbf{+IC} composes PIVOT with IndexCache. Boldface and underlining indicate
the best and second-best sparse methods in each column, respectively (DSA
excluded).}
\label{tab:longbench-ruler-combined}
\vspace{-0.5cm}
\end{table*}

\section{Experiments}

\subsection{Experimental Settings}

\subsubsection{Models and benchmarks.} We evaluate \method on two production-scale models that adopt the DSA indexer, DeepSeek-V3.2~\citep{deepseekv32} and GLM-5.1~\citep{glm5}. Both are served with MTP of depth $d=3$, emitting one decode token and three draft tokens per step, which fixes the decode group size to $g=d+1=4$. We report on two long-context suites: RULER~\citep{hsieh2024ruler}, which probes retrieval and reasoning over controllable lengths from $4$K to $128$K (20 samples per task-length pair), and LongBench~\citep{bai2024longbench}, which covers realistic long-document tasks across multiple domains. Together they test synthetic long-range recall and real-world long-context understanding.

\subsubsection{Baselines.} Since \method replaces only the indexer, we compare it against the original dense DSA indexer and against two orthogonal accelerators of that indexer: HISA (token axis) and IndexCache (layer axis). We also report \method composed with IndexCache (\textbf{+IC}).

\subsubsection{Implementation.} We serve all models with vLLM~\citep{kwon2023vllm} on NVIDIA H20 GPUs and use identical settings for all methods. \method has three hyperparameters: group size $g$, candidate budget $c$, and proxy aggregation strategy. We set $g=4$ by default in prefill and tie it to the MTP width in decoding, which is also $4$. We use $k=2048$, set $c=2k=4096$, and form the proxy by mean pooling (within a local window of size $w=4$, details can see in Appendix). \method-Reuse does not use a candidate set. We report sensitivity to $g$, $c$, and aggregation strategy in the ablation study (in prefill).

\subsection{Accuracy Results}
We evaluate both models on LongBench and RULER against the dense DSA indexer and two axis baselines, HISA (token) and IndexCache (layer). Table~\ref{tab:longbench-ruler-combined} reports the results.

\subsubsection{LongBench.} At these lengths, every sparse indexer stays within roughly half a point of the dense DSA baseline, so the benchmark does not stress the indexer and all methods preserve DSA-level quality. Within this narrow band, \method is on par with dense DSA on both models, with \method-Refine the stronger of the two variants (e.g., $56.18$ on DeepSeek-V3.2, just above the dense baseline). The subtask pattern is more informative than averages: \method gains most where queries are locally structured and a shared proxy summarizes them well (code completion, few-shot learning), and is slightly weaker where a query draws on dispersed evidence that a group-level proxy fits less tightly (multi-document QA). The two variants are nearly indistinguishable here, and composing with IndexCache is quality-neutral, consistent with the redundancy being real and the shared scan being lossless at these lengths.


\subsubsection{RULER.} RULER separates the methods only as the context grows: through moderate lengths all of them stay close to dense DSA, and the gaps open up at $64$K--$128$K. HISA degrades sharply here, $19$--$28$ points below dense at $128$K, as block-level pruning discards tokens later queries need. MISA stays closer through mid-lengths but still trails by $6$--$18$ at the extreme. \method-Refine, in contrast, tracks the dense baseline across all lengths, ending on par with both dense DSA and IndexCache. The two variants now diverge: \method-Reuse holds up through moderate lengths but declines faster at the extreme, because a group of queries attends to increasingly diverse content and a single shared selection loses recall that only per-query re-scoring can restore. This widening gap between Reuse and Refine is direct evidence that the fine step matters more as the context grows. Stacking IndexCache on top of \method remains lossless through moderate lengths and costs only a small margin at the longest contexts. Since \method alone already matches dense DSA, we treat \textbf{+IC} as an optional setting that trades this margin for additional speed.

\subsection{Efficiency Results}
\label{sec:efficiency}
Figure~\ref{fig:speedup} reports indexer speedups of \method over the dense DSA baseline, from short to very long context, for both the prefill and decode kernels and end-to-end.

\subsubsection{Kernel speedup.} Both variants speed up the indexer, and the gain grows with context length, because the shared full-prefix scan that \method amortizes is exactly the term that dominates as the prefix lengthens. At the longest context, the indexer kernel is several times faster (up to about $4.8\times$). The two variants differ by design at short context: \method-Reuse is faster throughout, while \method-Refine, which additionally re-scores candidates per query, is slower than dense DSA on short sequences and becomes favourable only once the prefix is long enough to amortize its extra work. This short-sequence overhead is exactly what our guardrail avoids: \method falls back to a dense scan for short groups, so it never runs slower than DSA in deployment.

\subsubsection{End-to-end speedup.} The kernel gains carry over to end-to-end latency, but only once the indexer accounts for a meaningful fraction of total inference: at short context the indexer is negligible and \method is essentially on par with the baseline, while at long context both variants become faster (reaching roughly $1.6\times$ end-to-end). This is consistent with the cost model in Method, in which the amortized $O(L^2)$ term dominates only as the context grows, so the speedup is largest exactly where long-context serving needs it most. Throughout, Reuse is the faster variant, and Refine trades part of that speed for the per-query accuracy, giving a clear speed--accuracy operating range between the two.

\subsection{Ablation Studies}
\label{sec:ablation}

The defaults of \method are guided by the observations rather than chosen purely by tuning. We ablate the four design choices on RULER (DeepSeek-V3.2; full two-model tables are provided in Appendix), and the results are consistent with the analysis behind our observations.

\begin{table}[ht]
\centering

\label{tab:ablation-proxy}

\begingroup
\small
\resizebox{\columnwidth}{!}{%
\begin{tabular}{@{}l !{\color{black!35}\vrule width 0.4pt} l *{6}{c}@{}}
\toprule
\multicolumn{1}{c}{} & & \multicolumn{5}{c}{\textbf{Context Length}} & \textbf{Overall} \\
\cmidrule(lr){3-7}\cmidrule(l){8-8}
\textbf{Variant} & \textbf{Pool}
& \textbf{8K} & \textbf{16K}
& \textbf{32K} & \textbf{64K} & \textbf{128K} & \textbf{AVG} \\
\midrule
\multirow{3}{*}{\textbf{PIVOT}-{\scriptsize\bfseries Reuse}}
 & Mean  & \textbf{95.99} & \textbf{95.48} & \textbf{95.30} & \textbf{88.71} & \textbf{87.96} & \textbf{92.69} \\
 & First & 94.37 & 89.13 & 89.56 & 85.22 & 79.56 & 87.57 \\
 & Last  & 92.81 & 92.98 & 90.11 & 85.71 & 82.33 & 88.79 \\
\cmidrule(lr){1-8}
\multirow{3}{*}{\textbf{PIVOT}-{\scriptsize\bfseries Refine}}
 & Mean  & \textbf{95.86} & \textbf{96.22} & \textbf{95.81} & \textbf{90.47} & \textbf{90.40} & \textbf{93.75} \\
 & First & 95.54 & 93.24 & 90.06 & 86.17 & 77.62 & 88.53 \\
 & Last  & 95.77 & 93.62 & 90.15 & 86.90 & 80.01 & 89.29 \\
\bottomrule
\end{tabular}%
}
\caption{\textbf{Proxy aggregation ablation on DeepSeek-V3.2 (RULER, \%).} Mean pooling vs. first/last query for \method-Reuse and \method-Refine, with $g=4$ and $c=4096$.}

\endgroup
\vspace{-0.5cm}
\end{table}

\subsubsection{Query aggregation.} Averaging the group into a proxy is far more robust than taking a single endpoint query: mean pooling stays close to dense DSA at all lengths, whereas an endpoint proxy tracks it only while the context fits the group's local span and then degrades sharply at long context (at $128$K the first-query proxy loses more than ten points to mean). A single query reflects only its own position and cannot represent the group as its members attend to increasingly diverse content, whereas the per-head mean captures the shared core identified by O1. We therefore adopt mean pooling.

\begin{table}[ht]
\centering

\label{tab:ablation-group}
\begingroup
\small
\setlength{\tabcolsep}{7.0pt}
\renewcommand{\arraystretch}{1.05}

\resizebox{\columnwidth}{!}{%
\begin{tabular}{@{}l c !{\color{black!35}\vrule width 0.4pt} *{6}{c}@{}}
\toprule
\multicolumn{2}{c}{} & \multicolumn{5}{c}{\textbf{Context Length}} & \textbf{Overall} \\
\cmidrule(lr){3-7}\cmidrule(l){8-8}
\textbf{Variant} & $\textbf{g}$
& \textbf{8K} & \textbf{16K}
& \textbf{32K} & \textbf{64K} & \textbf{128K} & \textbf{AVG} \\
\midrule
\multirow{4}{*}{\textbf{PIVOT}-{\scriptsize\bfseries Refine}}
& 4  & 95.86 & \textbf{96.22} & \textbf{95.81} & \textbf{90.47} & \textbf{90.40} & \textbf{93.75} \\
& 6  & 95.51 & 95.38 & 93.62 & 89.67 & 85.63 & 91.96 \\
& 8  & 95.96 & 95.10 & 94.34 & 87.44 & 82.90 & 91.15 \\
& 16 & \textbf{96.70} & 94.39 & 90.80 & 82.74 & 72.70 & 87.47 \\
\cmidrule(lr){1-8}
\multirow{4}{*}{\textbf{PIVOT}-{\scriptsize\bfseries Reuse}}
& 4  & \textbf{95.99} & \textbf{95.48} & \textbf{95.30} & \textbf{88.71} & \textbf{87.96} & \textbf{92.69} \\
& 6  & 95.45 & 95.19 & 93.10 & 87.22 & 83.69 & 90.93 \\
& 8  & 95.67 & 93.62 & 91.38 & 83.49 & 78.38 & 88.51 \\
& 16 & 95.38 & 91.41 & 83.88 & 76.84 & 65.16 & 82.53 \\
\bottomrule
\end{tabular}%
}
\endgroup
\caption{\textbf{Group size ablation on DeepSeek-V3.2 (RULER, \%).} Results use mean pooling and $c=4096$ during prefill.}
\vspace{-0.5cm}
\end{table}


\subsubsection{Group size $g$.} Accuracy decreases as the group size grows, and the loss concentrates at long context. This is the trade-off predicted by O2: a group shares a fixed candidate budget, but the union of its top-$k$ sets grows with $g$, so beyond a certain size the budget can no longer cover every member, and recall falls most where the union is largest. A small group such as $g{=}4$ keeps accuracy near-lossless while still amortizing the scan several-fold, which is why we adopt it.

\begin{table}[ht]
\centering

\label{tab:ablation-budget}

\begingroup
\small

\footnotesize
\setlength{\tabcolsep}{10pt}
\renewcommand{\arraystretch}{1.05}

\resizebox{1\columnwidth}{!}{%
\begin{tabular}{@{}l !{\color{black!35}\vrule width 0.4pt} *{6}{c}@{}}
\toprule
\multicolumn{1}{c}{} & \multicolumn{5}{c}{\textbf{Context Length}} & \textbf{Overall} \\
\cmidrule(lr){2-6}\cmidrule(l){7-7}
$\textbf{budget}$ & \textbf{8K} & \textbf{16K} & \textbf{32K} & \textbf{64K} & \textbf{128K} & \textbf{AVG} \\
\midrule
\textbf{3072} & \textbf{95.99} & 96.06 & 95.37 & 90.03 & 87.31 & 92.95 \\
\textbf{4096} & 95.86 & \textbf{96.22} & 95.81 & 90.47 & 90.40 & 93.75 \\
\textbf{6144} & 95.90 & 95.96 & \textbf{95.92} & 90.63 & 89.85 & 93.65 \\
\textbf{8192} & 95.51 & 95.58 & 95.54 & \textbf{92.24} & \textbf{90.55} & \textbf{93.88} \\
\bottomrule
\end{tabular}%
}
\endgroup
\caption{\textbf{Candidate budget ablation on DeepSeek-V3.2 (RULER, \%).} \method-Refine, mean pooling, $g=4$, prefill.}
\vspace{-0.5cm}
\end{table}


\subsubsection{Candidate budget $c$.} Accuracy saturates quickly in $c$: enlarging the budget beyond the default yields only marginal gains, whereas shrinking it well below the default starts to cost accuracy. This follows from O3: because the indexer scores are long-tailed, the relevant tokens for each query lie in a small high-scoring set that a modest budget already captures, so a larger budget mostly adds low-scoring tokens that do not change the top-$k$. Since a larger $c$ also raises the per-query re-scoring and sort cost, keeping $c$ near this knee maintains efficiency at no accuracy cost.

\begin{table}[!ht]
\centering

\label{tab:ablation-stage}

\begingroup
\small
\setlength{\tabcolsep}{10pt}
\renewcommand{\arraystretch}{1.05}

\resizebox{\columnwidth}{!}{%
\begin{tabular}{@{}l !{\color{black!35}\vrule width 0.4pt} *{6}{c}@{}}
\toprule
\multicolumn{1}{c}{} & \multicolumn{5}{c}{\textbf{Context Length}} & \textbf{Overall} \\
\cmidrule(lr){2-6}\cmidrule(l){7-7}
\textbf{Applied to} & \textbf{8K} & \textbf{16K} & \textbf{32K} & \textbf{64K} & \textbf{128K} & \textbf{AVG} \\
\midrule
\textbf{DSA}    & 95.71 & 96.12 & 95.77 & 91.32 & 90.45 & \textbf{93.87} \\
\textbf{Prefill only}   & \textbf{95.86} & \textbf{96.22} & 95.81 & 90.47 & 90.40 & 93.75 \\
\textbf{Decode only}    & 95.56 & 96.15 & 95.60 & \textbf{91.87} & 90.01 & 93.84 \\
\textbf{Both}           & \textbf{95.86} & 96.06 & \textbf{96.02} & 90.46 & \textbf{90.63} & 93.81 \\
\bottomrule
\end{tabular}%
}
\endgroup

\caption{\textbf{Deployment phase ablation on DeepSeek-V3.2 (RULER, \%).} \method-Refine with mean pooling, $g=4$, $c=4096$, ``Both'' applies \method to both prefill and decode.}
\vspace{-0.5cm}
\end{table}


\subsubsection{Where \method is applied.} \method is near-lossless whether enabled in prefill, in decode, or in both, all staying within a small margin of dense DSA. This confirms that a single algorithm serves both phases: it can be applied to either independently or to both jointly, allowing deployment to place the acceleration where it is most useful.
\section{Conclusion}

We presented \method, a group-shared indexer for token-level sparse attention. \method aggregates a group of queries into one proxy, runs a shared full-prefix scan to recall a candidate set, and lets each query take its top-$k$ from it. It offers two variants: the faster \method-Reuse, which shares the proxy's selection, and the default \method-Refine, which rescores per query to match the dense DSA indexer. One algorithm covers both inference stages, which differ only in grouping: by position in prefill, and by the tokens of one MTP step in decode. \method is training-free, leaves the Sparse MLA operator and KV cache unchanged, and is orthogonal to existing DSA accelerations. On DeepSeek-V3.2 and GLM-5.1 it retains dense-level accuracy on LongBench and RULER while accelerating the indexer by up to $4\times$, and reducing end-to-end latency by up to $1.6\times$ at long context. Combining the query axis with the token, head, and layer axes, and extending to broader model families, is promising future work.


\bibliography{aaai2027}
\appendix
\clearpage
\appendix
\onecolumn

\section{Appendix}

\subsection{A The DSA Indexer in Detail}
\label{app:dsa}
\subsubsection{Lightning indexer.}
We expand the indexer summarized in Section Preliminary. DSA augments each layer with a \emph{lightning indexer} that, for a query token $\mathbf{h}_t\!\in\!\mathbb{R}^{d}$ and a preceding token $\mathbf{h}_s\!\in\!\mathbb{R}^{d}$, computes an index score
\begin{equation}
I_{t,s} = \sum_{j=1}^{H^I} w^{I}_{t,j}\,\mathrm{ReLU}\!\big(\mathbf{q}^{I}_{t,j}\cdot \mathbf{k}^{I}_{s}\big),
\end{equation}
over $H^I$ indexer heads, where the indexing queries $\mathbf{q}^{I}_{t,j}\!\in\!\mathbb{R}^{d^I}$ and gating weights $w^{I}_{t,j}\!\in\!\mathbb{R}$ are linear projections of $\mathbf{h}_t$, and the indexing key $\mathbf{k}^{I}_{s}\!\in\!\mathbb{R}^{d^I}$ is a linear projection of $\mathbf{h}_s$. Two design choices keep the indexer lightweight: it uses far fewer heads and a smaller head dimension $d^I$ than the main attention, and the $\mathrm{ReLU}$ activation is chosen for throughput over a softmax-style normalizer. Together these let the indexer be implemented in FP8, so that---despite scanning the full prefix---its per-score cost is small relative to the main attention.

\subsubsection{Fine-grained token selection.}
Given the scores $\{I_{t,s}\}_{s\le t}$, the fine-grained token-selection mechanism retrieves the key--value entries $\{\mathbf{c}_s\}$ of the $k$ highest-scoring tokens and computes the attention output over only that subset,
\begin{equation}
\begin{aligned}
\mathbf{u}_t
&= \mathrm{Attn}\!\big(
\mathbf{h}_t,\ 
\{\mathbf{c}_s \mid I_{t,s}\in \mathrm{Top\text{-}}k(I_{t,:})\}
\big), \\
\mathcal{T}_t
&= \{s : I_{t,s}\in \mathrm{Top\text{-}}k(I_{t,:})\}.
\end{aligned}
\end{equation}

Writing $\mathcal{T}_t$ for this selected index set, the indexer and the Sparse MLA operator communicate through $\mathcal{T}_t$ alone. This makes $\mathcal{T}_t$ the sole interface \method must preserve: it changes only how $\mathcal{T}_t$ is produced and leaves the operator, the selection size $k$, and the KV cache unchanged. Setting the group size to one recovers the original per-query indexer exactly.
\subsubsection{Instantiation under MLA.}
For continued training from a Multi-head Latent Attention (MLA) backbone, DSA instantiates the selected attention in the MQA mode of MLA: each latent key--value entry $\mathbf{c}_s$ is shared across all query heads of token $t$, so the top-$k$ selection is made once per token rather than once per head. This is what makes the token-level selection $\mathcal{T}_t$ well defined at the token granularity that \method groups over.

In the FP8 kernel implementation, the score \eqref{eq:indexer} is realized as
\begin{equation}
I_{t,s} = \Big(\textstyle\sum_{j} w_{t,j}\,\mathrm{ReLU}\big(\mathbf{q}^{\text{fp8}}_{t,j}\cdot \mathbf{k}^{\text{fp8}}_{s}\big)\Big)\cdot k^{\text{scale}}_{s},
\end{equation}
where the per-token key scale $k^{\text{scale}}_{s}$ restores the FP8-quantized magnitudes and the gating weights $w_{t,j}$ absorb the constant factor $q^{\text{scale}}\cdot \texttt{softmax\_scale}\cdot (H^{I})^{-1/2}$. The dense logits are materialized in chunks: the fused MQA-logits kernel (\texttt{fp8\_fp4\_mqa\_logits}) is subject to a $512$\,MB workspace bound, so long prefixes are tiled along the key axis. 

\subsection{B The Local Window}
\label{app:window}

\subsubsection{Why a local window.} The shared candidate set $\mathcal{C}$ is recalled by a single proxy scan, and a query's most recent tokens (those closest to its own position) are the ones most likely to be selected yet also the ones a group-level proxy represents least well, especially for later members of the group whose recent context was not yet present when $\mathcal{C}$ was formed. \method therefore augments $\mathcal{C}$ with a per-query local window $W_t=[\,t-W+1,\ t\,]$, the $W$ most recent tokens up to position $t$ (including $t$ itself), and refines each query over $\mathcal{C}\cup W_t$. This guarantees that every query retains its local context regardless of the proxy, at negligible cost since $W$ is small.

\subsubsection{Decode.} Each decode step builds its own candidate pool, so the window is applied per step with width $W$: with a group of $g=d{+}1$ queries at consecutive positions, we require $W\ge g$ so that the window covers every token generated within the step (which are absent from the pool recalled at the step's first position). Each query then refines over (pool $\cup$ its own $W_t$), with duplicates removed.

\subsubsection{Prefill.} In prefill the $g$ queries of a group \emph{share} one candidate set, so their windows must all fit inside it. The per-query windows $[\,t-W+1,\ t\,]$ of $g$ consecutive positions are staggered, and their union spans $W+g-1$ tokens. To keep the total candidate budget fixed at $c$, we reserve $w_g=W+g-1$ slots for this union and let the proxy fill the remaining top-$(c-w_g)$; the two parts concatenate to exactly $c$ candidates, so the refine width does not grow with $W$. Using $w_g=W+g-1$ (rather than $W$) ensures each row obtains its full $W$-token window, matching the per-step behavior of decode; reserving only $W$ would drop up to $g-1$ of the oldest window tokens for the earliest rows of the group. Setting $W=0$ recovers the plain top-$c$ candidate set.

\subsection{C Per-layer analysis of query locality in indexer top-$k$ selection}
\label{app:obs-study}

\begin{figure*}[ht]
    \centering
    \includegraphics[width=1.0\linewidth]{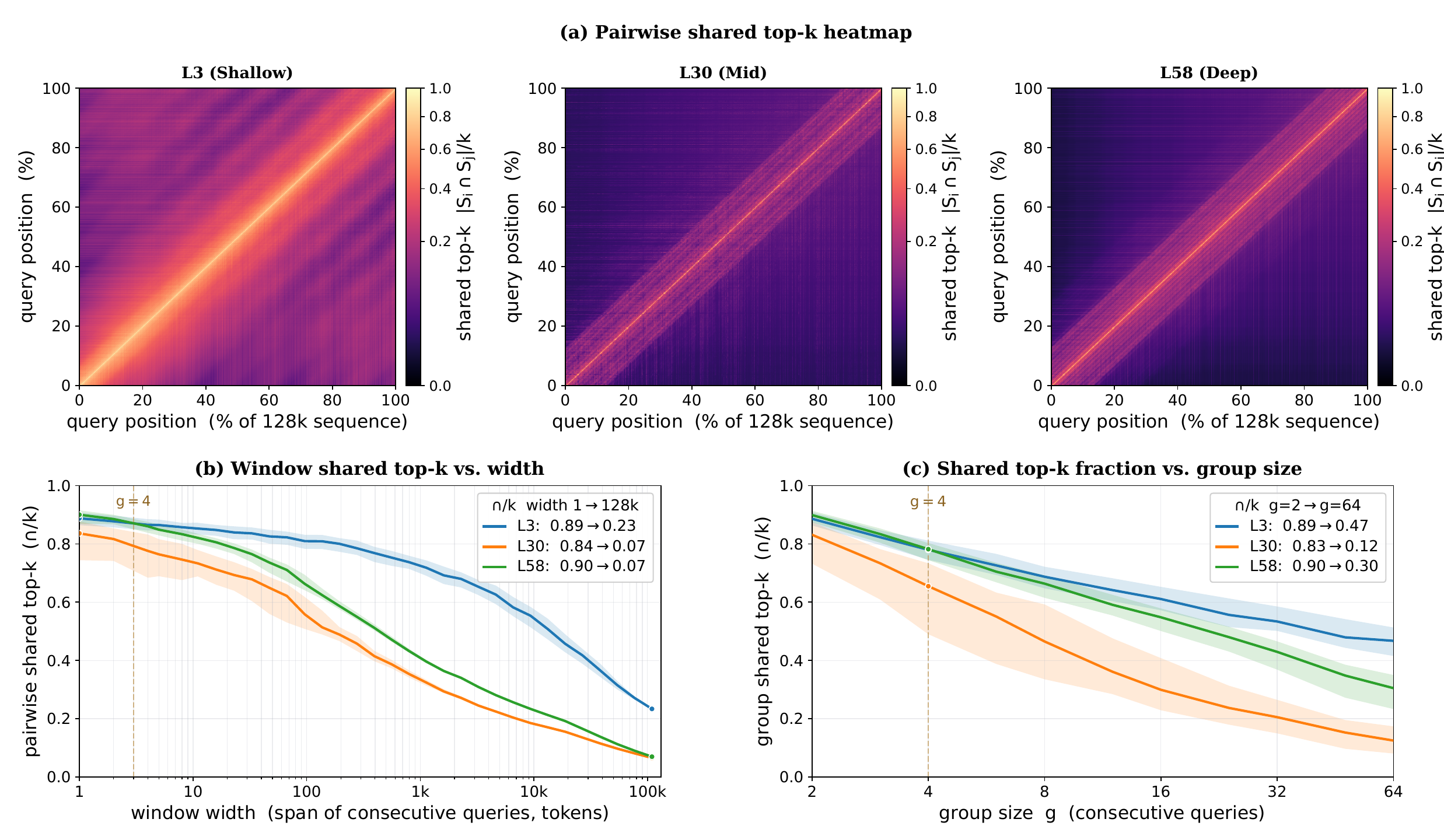}
    \caption{\textbf{Per-layer view of query locality in indexer top-$k$ selection}
    (appendix to Fig.~\ref{fig:obs} DeepSeek-V3.2, RULER-QA 128K; deep
    queries ($k_e\!>\!20$K), budget $k\!=\!2048$; layers L3 (shallow), L30 (mid),
    L58 (deep). \textbf{(a)~Shared-key heatmaps.} Pairwise shared-key fraction
    $|S_i\!\cap\!S_j|/k$ between query positions. Every layer shows a bright diagonal
    band---neighbouring queries select almost the same top-$k$ keys---widest at the
    shallow layer and progressively sharper with depth. \textbf{(b)~vs.\ window
    width.} The mean pairwise $|S_i\!\cap\!S_j|/k$ over a window of consecutive
    queries decays with the window span: adjacent queries share $\approx\!0.84$--$0.90$
    (width~1) and only $0.07$--$0.23$ at the full 128K span. \textbf{(c)~vs.\ group
    size.} The joint shared fraction $|\bigcap_i \mathrm{top}\text{-}k_i|/k$ over a
    group of $g$ consecutive queries decreases with $g$ but stays substantial for
    small groups ($0.83$--$0.90$ at $g\!=\!2$; $0.12$--$0.47$ at $g\!=\!64$); the dashed
    line marks the $g\!=\!4$ operating point. Together these confirm the query
    locality exploited by Grouped-Proxy holds from shallow to deep layers.}
    \label{fig:appendix_c1}
\end{figure*}

Observation O1 states that neighbouring queries select nearly the same top-$k$ keys, and Figure~\ref{fig:obs} summarizes it across the model. Here we verify that this locality is not an artifact of any single layer but holds throughout the network, and we quantify how it decays with query distance and group size, which is what determines a safe operating point for the group size $g$. We measure it on DeepSeek-V3.2 over RULER-QA at 128K, on deep queries ($k_e\!>\!20$K) with budget $k\!=\!2048$, at a shallow, a middle, and a deep layer (L3, L30, L58); Figure~\ref{fig:appendix_c1} reports three views.

The pairwise shared-key heatmaps (a) show a bright diagonal band at every layer: adjacent queries overlap heavily in their selected keys, and the band, if anything, sharpens with depth. Quantifying this decay (b), the shared fraction falls from $0.84$--$0.90$ for immediate neighbours to $0.07$--$0.23$ across the full 128K span, so locality is strong locally but does not extend arbitrarily far, matching the group-shared design rather than a global one. Finally, the joint shared fraction over a group of $g$ consecutive queries (c) decreases gracefully with $g$ and remains high for small groups ($0.83$--$0.90$ at $g\!=\!2$), degrading only for large $g$ ($0.12$--$0.47$ at $g\!=\!64$); the default $g\!=\!4$ sits well inside the high-overlap regime. Together, these confirm that the query locality \method exploits holds from shallow to deep layers and justify a moderate group size.

\vspace{-0.5cm}
\subsection{D Indexer kernel latency breakdown }

\begin{figure*}[!htbp]
    \centering
    \includegraphics[width=1.0\linewidth]{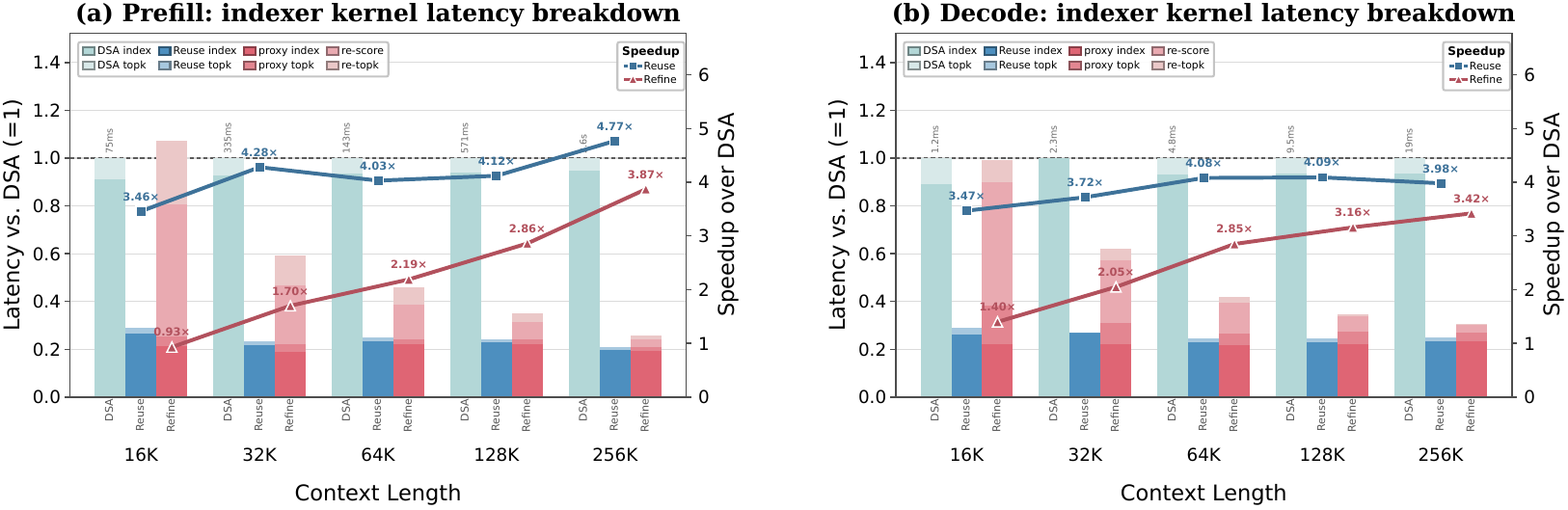}
    \caption{\textbf{Indexer kernel latency breakdown and speedup over the DSA baseline on DeepSeek-V3.2}, from 16K to 256K. \textbf{(a)} Prefill and \textbf{(b)} decode. Stacked bars show the latency of individual kernel stages normalized to DSA ($=1$), with absolute latency annotated above each bar. Lines report the corresponding speedup over DSA on the right axis. PIVOT-Reuse consistently achieves around $4\times$ speedup, while PIVOT-Refine delivers increasing speedup as the context length grows.}
    \label{fig:indexer_breakdown}
\end{figure*}

We profile the indexer kernels on an NVIDIA H20 GPU. As shown in Fig.~\ref{fig:indexer_breakdown}, index computation remains the dominant cost for both DSA and PIVOT-Reuse. For PIVOT-Refine, the dominant component shifts from refinement to proxy construction as the context length increases: re-score and re-top-$k$ account for 76\% (prefill) and 85\% (decode) of its latency at 16K, whereas proxy index and proxy top-$k$ account for 81\% and 92\%, respectively, at 256K. This shift indicates that refinement overhead grows slowly and is progressively amortized at longer contexts. Consequently, PIVOT-Refine improves from $0.93\times$/$1.40\times$ speedup at 16K to $3.87\times$/$3.42\times$ at 256K for prefill/decode, while PIVOT-Reuse consistently achieves approximately $4\times$ speedup.

\vspace{-0.5cm}

\subsection{E Algorithm}
\definecolor{bluephase}{RGB}{30,90,180}
\definecolor{redphase}{RGB}{180,50,50}
\definecolor{orangevar}{RGB}{210,120,20}
\definecolor{tealvar}{RGB}{20,140,130}

\begin{algorithm}[!ht]
\caption{\method Inference}

\label{alg:pivot}
\begin{algorithmic}[1]
\Require indexing queries $\{\mathbf{q}^{I}_{t,j}\}$, gating weights $\{w^{I}_{t,j}\}$, indexing keys $\{\mathbf{k}^{I}_s\}_{s=1}^{L}$; group size $g$; candidate budget $c$; window $W$; token budget $k$; variant $v$
\Ensure selected token set $\mathcal{T}_t$ of size $k$ for every query $t$

\Statex \textbf{Stage 0: form query groups}
\If{\textcolor{bluephase}{prefill}}
  \State \textcolor{bluephase}{Partition positions into fixed-size groups $G=\{t_0,\dots,t_0{+}g{-}1\}$}
\ElsIf{\textcolor{redphase}{decode}}
  \State \textcolor{redphase}{Group the current token with its $d$ MTP draft tokens, $g=d{+}1$}
\EndIf

\Statex
\ForAll{groups $G$ with first position $t_0=\min G$}
  \If{$t_0 < c$}
    \State Densely score each $t\in G$ over $[1,t]$ and take its top-$k$
    \State \textbf{continue} \Comment{short-group guardrail}
  \EndIf

  \Statex \textbf{Stage 1: shared proxy scan (coarse)}
  \State $\bar{\mathbf{q}}^{I}_{j}\gets \frac{1}{g}\sum_{t\in G}\mathbf{q}^{I}_{t,j}$,\quad
         $\bar{w}^{I}_{j}\gets \frac{1}{g}\sum_{t\in G} w^{I}_{t,j}$ \Comment{per-head mean proxy}

  \State $\bar{I}_{s}\gets \sum_{j} \bar{w}^{I}_{j}\,
  \mathrm{ReLU}\!\left(\bar{\mathbf{q}}^{I}_{j}\cdot\mathbf{k}^{I}_{s}\right)$ for $s=1,\dots,t_0$ \Comment{one full-prefix scan}

  \Statex \textbf{Stage 2: per-query selection (fine)}
  \If{$v=\textcolor{orangevar}{\textsc{Reuse}}$}
    \State \textcolor{orangevar}{$\mathcal{T}_t\gets \mathrm{TopK}(\{\bar{I}_{s}\},\,k)$ for all $t\in G$} \Comment{\textcolor{orangevar}{whole group shares the proxy top-$k$}}

  \ElsIf{$v=\textcolor{tealvar}{\textsc{Refine}}$}
    \State \textcolor{tealvar}{$\mathcal{C}\gets \mathrm{TopK}(\{\bar{I}_{s}\},\,c)$} \Comment{\textcolor{tealvar}{shared candidate set}}

    \ForAll{$t\in G$}
      \State \textcolor{tealvar}{$I_{t,s}\gets \sum_{j} w^{I}_{t,j}\,
      \mathrm{ReLU}\!\left(\mathbf{q}^{I}_{t,j}\cdot\mathbf{k}^{I}_{s}\right)$ for $s\in\mathcal{C}$} \Comment{\textcolor{tealvar}{exact per-query re-score}}

      \State \textcolor{tealvar}{$\mathcal{T}_t\gets \mathrm{TopK}(\{I_{t,s}\mid s\in\mathcal{C}\},\,k)$} \Comment{\textcolor{tealvar}{query-specific top-$k$}}
    \EndFor
  \EndIf
\EndFor

\State \textbf{return} $\{\mathcal{T}_t\}$ to Sparse MLA through the unchanged DSA interface
\end{algorithmic}
\end{algorithm}

\clearpage
\subsection{F Full Ablations}
\label{app:ablation}

We report the complete two-model ablations that Section Ablation. summarizes on DeepSeek-V3.2. All runs are on RULER from 4K to 128K, and we report overall accuracy (\%). Except for the deployment-stage study of Table~\ref{tab:abl-stage}, which by design varies the phase in which \method is enabled, all ablations are conducted in the prefill phase. Table~\ref{tab:abl-pool} ablates the query aggregation, Table~\ref{tab:abl-group} the group size $g$, Table~\ref{tab:abl-budget} the candidate budget $c$, and Table~\ref{tab:abl-stage} the deployment stage. Best per column in bold. Unless varied, we use \method-Refine with mean pooling, $g=4$, and $c=4096$; the trends hold on both DeepSeek-V3.2 and GLM-5.1.

\begin{table}[ht]
\centering

\begingroup
\small
\setlength{\tabcolsep}{12.0pt}
\renewcommand{\arraystretch}{1.0}

\resizebox{0.6\columnwidth}{!}{%
\begin{tabular}{@{}l !{\color{black!35}\vrule width 0.4pt} l *{7}{c}@{}}
\toprule
\multicolumn{1}{c}{} & & \multicolumn{6}{c}{\textbf{Context Length}} & \textbf{Overall} \\
\cmidrule(lr){2-8}\cmidrule(l){9-9}
\textbf{Variant} & \textbf{Pool}
& \textbf{4K} & \textbf{8K} & \textbf{16K}
& \textbf{32K} & \textbf{64K} & \textbf{128K} & \textbf{AVG} \\
\midrule
\rowcolor{tblecho}
\multicolumn{9}{c}{\rule{0pt}{2.3ex}\textbf{DeepSeek-V3.2}} \\
\midrule
\multirow{3}{*}{\textbf{PIVOT}-{\scriptsize\bfseries Reuse}}
 & Mean  & 96.03 & \textbf{95.99} & \textbf{95.48} & \textbf{95.30} & \textbf{88.71} & \textbf{87.96} & \textbf{93.25} \\
 & First & \textbf{96.41} & 94.37 & 89.13 & 89.56 & 85.22 & 79.56 & 89.04 \\
 & Last  & \textbf{96.41} & 92.81 & 92.98 & 90.11 & 85.71 & 82.33 & 90.06 \\
\cmidrule(lr){1-9}
\multirow{3}{*}{\textbf{PIVOT}-{\scriptsize\bfseries Refine}}
 & Mean  & 96.03 & \textbf{95.86} & \textbf{96.22} & \textbf{95.81} & \textbf{90.47} & \textbf{90.40} & \textbf{94.13} \\
 & First & \textbf{96.41} & 95.54 & 93.24 & 90.06 & 86.17 & 77.62 & 89.84 \\
 & Last  & \textbf{96.41} & 95.77 & 93.62 & 90.15 & 86.90 & 80.01 & 90.48 \\
\midrule
\rowcolor{tblecho}
\multicolumn{9}{c}{\rule{0pt}{2.3ex}\textbf{GLM-5.1}} \\
\midrule
\multirow{3}{*}{\textbf{PIVOT}-{\scriptsize\bfseries Reuse}}
 & Mean  & \textbf{94.74} & \textbf{95.51} & \textbf{96.15} & \textbf{95.52} & \textbf{92.56} & \textbf{89.83} & \textbf{94.05} \\
 & First & 93.08 & 94.49 & 93.81 & 89.73 & 89.26 & 83.62 & 90.67 \\
 & Last  & 93.85 & 91.54 & 91.06 & 89.93 & 84.78 & 85.63 & 89.47 \\
\cmidrule(lr){1-9}
\multirow{3}{*}{\textbf{PIVOT}-{\scriptsize\bfseries Refine}}
 & Mean  & \textbf{95.90} & 95.77 & \textbf{96.08} & \textbf{96.37} & \textbf{94.73} & \textbf{91.83} & \textbf{95.11} \\
 & First & 94.49 & 95.77 & 93.00 & 90.64 & 88.13 & 85.67 & 91.28 \\
 & Last  & 95.13 & \textbf{96.03} & 94.54 & 92.87 & 89.85 & 86.76 & 92.53 \\
\bottomrule
\end{tabular}%
}

\endgroup

\caption{\textbf{Ablation on query aggregation (RULER, \%).} mean vs.\ first vs.\ last, under both modes at $g=4$, $c=4096$. Best per column in bold.}
\label{tab:abl-pool}
\end{table}

\begin{table}[ht]
\centering
\vspace{-0.5cm}

\begingroup
\small
\setlength{\tabcolsep}{15.0pt}
\renewcommand{\arraystretch}{1.0}

\resizebox{0.6\columnwidth}{!}{%
\begin{tabular}{@{}l !{\color{black!35}\vrule width 0.4pt} *{7}{c}@{}}
\toprule
\multicolumn{1}{c}{} & \multicolumn{6}{c}{\textbf{Context Length}} & \textbf{Overall} \\
\cmidrule(lr){2-7}\cmidrule(l){8-8}
\textbf{budget} $\textbf{c}$ & \textbf{4K} & \textbf{8K} & \textbf{16K} & \textbf{32K} & \textbf{64K} & \textbf{128K} & \textbf{AVG} \\
\midrule
\rowcolor{tblecho}
\multicolumn{8}{c}{\rule{0pt}{2.3ex}\textbf{DeepSeek-V3.2}} \\
\midrule
\textbf{3072} & 96.03 & \textbf{95.99} & 96.06 & 95.37 & 90.03 & 87.31 & 93.46 \\
\textbf{4096} & 96.03 & 95.86 & \textbf{96.22} & 95.81 & 90.47 & 90.40 & 94.13 \\
\textbf{6144} & \textbf{96.41} & 95.90 & 95.96 & \textbf{95.92} & 90.63 & 89.85 & 94.11 \\
\textbf{8192} & 96.03 & 95.51 & 95.58 & 95.54 & \textbf{92.24} & \textbf{90.55} & \textbf{94.24} \\
\midrule
\rowcolor{tblecho}
\multicolumn{8}{c}{\rule{0pt}{2.3ex}\textbf{GLM-5.1}} \\
\midrule
\textbf{3072} & 94.62 & 95.77 & \textbf{96.15} & 96.08 & 93.46 & 91.53 & 94.60 \\
\textbf{4096} & \textbf{95.90} & 95.77 & 96.08 & 96.37 & 94.73 & 91.83 & 95.11 \\
\textbf{6144} & 95.38 & \textbf{96.41} & 95.69 & \textbf{96.50} & 94.92 & \textbf{92.60} & \textbf{95.25} \\
\textbf{8192} & 95.51 & 96.15 & 95.77 & 96.38 & \textbf{95.28} & 91.83 & 95.16 \\
\bottomrule
\end{tabular}%
}
\endgroup
\caption{\textbf{Ablation on candidate budget $c$ (RULER, \%).} PIVOT-Refine with mean pooling at $g=4$ during prefill. Best per column in bold.}

\label{tab:abl-budget}
\vspace{-0.5cm}
\end{table}

\begin{table}[!ht]
\centering

\begingroup
\small
\setlength{\tabcolsep}{12.0pt}
\renewcommand{\arraystretch}{1.0}

\resizebox{0.6\columnwidth}{!}{%
\begin{tabular}{@{}l c !{\color{black!35}\vrule width 0.4pt} *{7}{c}@{}}
\toprule
\multicolumn{2}{c}{} & \multicolumn{6}{c}{\textbf{Context Length}} & \textbf{Overall} \\
\cmidrule(lr){3-8}\cmidrule(l){9-9}
\textbf{Variant} & $g$
& \textbf{4K} & \textbf{8K} & \textbf{16K}
& \textbf{32K} & \textbf{64K} & \textbf{128K} & \textbf{AVG} \\
\midrule
\rowcolor{tblecho}
\multicolumn{9}{c}{\rule{0pt}{2.3ex}\textbf{DeepSeek-V3.2}} \\
\midrule
\multirow{4}{*}{\textbf{PIVOT}-{\scriptsize\bfseries Refine}}
& 4  & 96.03 & 95.86 & \textbf{96.22} & \textbf{95.81} & \textbf{90.47} & \textbf{90.40} & \textbf{94.13} \\
& 6  & 96.41 & 95.51 & 95.38 & 93.62 & 89.67 & 85.63 & 92.71 \\
& 8  & \textbf{96.79} & 95.96 & 95.10 & 94.34 & 87.44 & 82.90 & 92.09 \\
& 16 & 95.64 & \textbf{96.70} & 94.39 & 90.80 & 82.74 & 72.70 & 88.83 \\
\cmidrule(lr){1-9}
\multirow{4}{*}{\textbf{PIVOT}-{\scriptsize\bfseries Reuse}}
& 4  & \textbf{96.03} & \textbf{95.99} & \textbf{95.48} & \textbf{95.30} & \textbf{88.71} & \textbf{87.96} & \textbf{93.25} \\
& 6  & 95.64 & 95.45 & 95.19 & 93.10 & 87.22 & 83.69 & 91.72 \\
& 8  & \textbf{96.03} & 95.67 & 93.62 & 91.38 & 83.49 & 78.38 & 89.76 \\
& 16 & \textbf{96.03} & 95.38 & 91.41 & 83.88 & 76.84 & 65.16 & 84.78 \\
\midrule
\rowcolor{tblecho}
\multicolumn{9}{c}{\rule{0pt}{2.3ex}\textbf{GLM-5.1}} \\
\midrule
\multirow{4}{*}{\textbf{PIVOT}-{\scriptsize\bfseries Refine}}
& 4  & \textbf{95.90} & 95.77 & \textbf{96.08} & 96.37 & \textbf{94.73} & \textbf{91.83} & \textbf{95.11} \\
& 6  & 95.51 & 96.03 & \textbf{96.08} & \textbf{96.42} & 92.00 & 88.33 & 94.06 \\
& 8  & 95.38 & \textbf{96.41} & 95.69 & 94.79 & 89.99 & 85.85 & 93.02 \\
& 16 & 95.51 & 95.38 & 95.18 & 91.00 & 87.16 & 82.40 & 91.11 \\
\cmidrule(lr){1-9}
\multirow{4}{*}{\textbf{PIVOT}-{\scriptsize\bfseries Reuse}}
& 4  & 94.74 & 95.51 & \textbf{96.15} & \textbf{95.52} & \textbf{92.56} & \textbf{89.83} & \textbf{94.05} \\
& 6  & 94.62 & 95.64 & 95.69 & 95.18 & 89.67 & 84.04 & 92.47 \\
& 8  & \textbf{95.00} & \textbf{96.41} & 94.92 & 92.26 & 88.01 & 82.35 & 91.49 \\
& 16 & 94.36 & 94.49 & 90.31 & 86.74 & 80.63 & 79.22 & 87.63 \\
\bottomrule
\end{tabular}%
}
\endgroup

\caption{\textbf{Ablation on group size $g$ (RULER, \%).} We compare PIVOT-Refine and PIVOT-Reuse with mean pooling and $c=4096$ during prefill. Best per column within each model and variant in bold.}

\label{tab:abl-group}
\vspace{-0.5cm}
\end{table}

\clearpage
\begin{table}[!htbp]
\centering

\begingroup
\small
\setlength{\tabcolsep}{12.0pt}
\renewcommand{\arraystretch}{1.0}

\resizebox{0.6\columnwidth}{!}{%
\begin{tabular}{@{}l l !{\color{black!35}\vrule width 0.4pt} *{7}{c}@{}}
\toprule
\multicolumn{2}{c}{} & \multicolumn{6}{c}{\textbf{Context Length}} & \textbf{Overall} \\
\cmidrule(lr){3-8}\cmidrule(l){9-9}
\textbf{Variant} & \textbf{Applied to}
& \textbf{4K} & \textbf{8K} & \textbf{16K}
& \textbf{32K} & \textbf{64K} & \textbf{128K} & \textbf{AVG} \\
\midrule
\rowcolor{tblecho}
\multicolumn{9}{c}{\rule{0pt}{2.3ex}\textbf{DeepSeek-V3.2}} \\
\midrule
\multirow{4}{*}{\textbf{PIVOT}-{\scriptsize\bfseries Refine}}
& \textbf{DSA}          & \textbf{96.41} & 95.71 & 96.12 & 95.77 & 91.32 & 90.45 & \textbf{94.30} \\
& \textbf{Prefill only} & 96.03 & \textbf{95.86} & \textbf{96.22} & 95.81 & 90.47 & 90.40 & 94.13 \\
& \textbf{Decode only}  & \textbf{96.41} & 95.56 & 96.15 & 95.60 & \textbf{91.87} & 90.01 & 94.27 \\
& \textbf{Both}         & \textbf{96.41} & \textbf{95.86} & 96.06 & \textbf{96.02} & 90.46 & \textbf{90.63} & 94.24 \\
\cmidrule(lr){1-9}
\multirow{4}{*}{\textbf{PIVOT}-{\scriptsize\bfseries Reuse}}
& \textbf{DSA}          & \textbf{96.41} & 95.71 & \textbf{96.12} & \textbf{95.77} & 91.32 & \textbf{90.45} & \textbf{94.30} \\
& \textbf{Prefill only} & 96.03 & \textbf{95.99} & 95.48 & 95.30 & 88.71 & 87.96 & 93.25 \\
& \textbf{Decode only}  & 95.64 & 95.67 & 96.06 & 95.74 & \textbf{92.31} & 88.58 & 94.00 \\
& \textbf{Both}         & 96.03 & 95.86 & 94.81 & 95.16 & 90.97 & 86.46 & 93.22 \\
\midrule
\rowcolor{tblecho}
\multicolumn{9}{c}{\rule{0pt}{2.3ex}\textbf{GLM-5.1}} \\
\midrule
\multirow{4}{*}{\textbf{PIVOT}-{\scriptsize\bfseries Refine}}
& \textbf{DSA}          & 95.51 & \textbf{96.15} & 96.03 & 96.03 & 95.27 & 92.14 & 95.19 \\
& \textbf{Prefill only} & \textbf{95.90} & 95.77 & 96.08 & 96.37 & 94.73 & 91.83 & 95.11 \\
& \textbf{Decode only}  & 95.00 & \textbf{96.15} & \textbf{96.15} & 96.08 & \textbf{95.74} & \textbf{92.79} & \textbf{95.32} \\
& \textbf{Both}         & 95.42 & \textbf{96.15} & \textbf{96.15} & \textbf{96.46} & 95.03 & 91.79 & 95.17 \\
\cmidrule(lr){1-9}
\multirow{4}{*}{\textbf{PIVOT}-{\scriptsize\bfseries Reuse}}
& \textbf{DSA}          & \textbf{95.51} & \textbf{96.15} & 96.03 & 96.03 & \textbf{95.27} & \textbf{92.14} & \textbf{95.19} \\
& \textbf{Prefill only} & 94.74 & 95.51 & 96.15 & 95.52 & 92.56 & 89.83 & 94.05 \\
& \textbf{Decode only}  & 95.00 & 95.26 & \textbf{96.54} & \textbf{96.15} & 95.23 & 91.51 & 94.95 \\
& \textbf{Both}         & 94.62 & \textbf{96.15} & 96.06 & 95.62 & 93.59 & 88.86 & 94.15 \\
\bottomrule
\end{tabular}%
}
\endgroup

\caption{\textbf{Ablation on PIVOT deployment stage (RULER, \%).} PIVOT-Refine and PIVOT-Reuse with mean pooling at $g=4$ and $c=4096$; ``Both'' enables PIVOT during prefill and decode. Best per column within each model and variant in bold.}

\label{tab:abl-stage}
\end{table}

\end{document}